# A Novel Approach for Stable Selection of Informative Redundant Features from High Dimensional fMRI Data

Yilun Wang, Zhiqiang Li ,Yifeng Wang, Xiaona Wang, Junjie Zheng, Xujun Duan and Huafu Chen

*Abstract*— Recognition of different cognitive states or detection of mental disorders from function Magnetic Resonance Imaging (fMRI) has attracted increasing attention. It can often be formulated as a pattern recognition problem to classify different cognitive states or discriminate patients of mental disorders from normal controls. Feature selection is among the most important components because it not only helps enhance the classification accuracy, but also or even more important provides potential biomarker discovery. However, traditional multivariate methods is likely to obtain unstable and unreliable results in case of an extremely high dimensional feature space and very limited training samples, where the features are often correlated or redundant. In order to improve the stability, generalization and interpretations of the discovered potential biomarker and enhance the robustness of the resultant classifier, the redundant but informative features need to be also selected. Therefore we introduced a novel feature selection method which combines a recent implementation of the stability selection approach and the elastic net approach. Compared with the traditional stability selection implementation with the pure $\ell_1$-norm regularized model serving as the baseline model, the proposed method achieves a better empirical sensitivity, because elastic net encourages a grouping effect besides sparsity, thus making use of the feature redundancy. Compared with the feature selection method based on the plain elastic net, our method achieves the finite sample control for certain error rates of false discoveries, transparent principle for choosing a proper amount of regularization and the robustness of the feature selection results, due to the incorporation of stability selection. The advantage in terms of better control of false discoveries and missed discoveries of our approach, and the resulted better interpretability of the obtained potential biomarker is verified in both synthetic and real fMRI experiments. In addition, we are among the first to demonstrate the "robustness" of feature selection benefiting from the incorporation of stability selection and also among the first to demonstrate the possible "unrobustness" of the classical univariate two-sample t-test method. Specifically, we show the "robustness" of our feature selection results in existence of noisy (wrong) training labels, as well as the "robustness" of the resulted classifier based on our feature selection results in the existence of data variation, demonstrated by a multi-center attention-deficit/hyperactivity disorder (ADHD) fMRI data, which is of big variation across different centers.

## I. INTRODUCTION

### A. Our research motivation and backgrounds

One objective of functional Magnetic Resonance Imaging (fMRI) to measure and map brain activity in a task, a process or an emotion, in a noninvasive and safe means. Numerous functional imaging studies have reported neural activities during the experience of specific emotions or cognitive activities and demonstrated the potentials of functional imaging MRI for the classification of cognitive states or identification of mental disorders. In this paper, we consider learning from fMRI data as a pattern recognition problem and mainly focus on how to accurately and stably identify the relevant features (either voxels or network connections) that participate in a given cognitive task or that are closely related with certain mental disorders. In this paper, we will mainly consider the binary classification problems such as discriminating patients of certain mental disorder from the normal persons or classifying different cognative states, though the proposed idea can also be extended to the case of regression

As we know, with the rapid development of data capture and storage technologies, the "curse of dimensionality" becomes a common issue in many fields [1] including the field of pattern recognition and machine learning, where "curse of dimensionality" often refers to an extremely high dimensional feature space. Therefore, feature selection, as a way of dimensional reduction, is critical in many pattern recognition applications such as medical image analysis, computer vision, speech recognition and many more [2]. In this paper, we consider the related challeges in the neuroimaging data based pattern recognition, where besides the "curse of dimensionality", feature selection has another common difficulty, which lies in the small number of training samples, due to varied reasons. This "high dimensionality and small-sample-size" issue, where the number of features or variables is overwhelmingly larger than the sample size, is an important distinction between neural imaging data analysis with many other kinds of data analysis problems such as big data analysis from social media [3]. More important, the need of feature selection in neuroimaging data analysis is often inspired by two lines of evidence. One is that the selected features are used to establish a classifier of better classification or prediction accuracy [4]. It can also make the predictors faster and more cost-effective, and provides valuable information for better understanding of the underlying processing that generated the data. The other one is that the selected features are often considered as the potential biomarker for medical diagnosis or may shed light on biological processing involved in various diseases and suggest novel targets [5]. Therefore, a relatively restrict control of both false positive and false negatives when selecting the discriminative features is strongly required. Thus, how to perform accurate and stable feature selection has becomes a challenge and also an important problem in neuroimaging analysis.

Yilun Wang is with School of Mathematical Sciences and Center for Information in Biomedicine, University of Electronic Science and Technology of China, Chengdu Sichuan, 611731 P. R. China. (e-mail: yilun.wang@rice.edu). He is also with Center for Applied Mathematics, Cornell University, Ithaca, NY 14853 USA

Zhiqiang Li, Yifeng Wang, Xiaona Wang, Junjie Zheng, Xujun Duan and Huafu Chen are with Key laboratory for Neuroinformation of Ministry of Education, School of Life Science and Technology, and Center for Information in Biomedicine, University of Electronic Science and Technology of China, Chengdu, Sichuan, 611054, P. R. China. (e-mail: Chenhf@uestc.edu.cn)

Yilun Wang and Zhiqiang Li contributed equally to this work.

Corresponding authors: Huafu Chen

To overcome the common challege that the number of samples of medical imaging data is generally much smaller than the dimension of the feature space, we focus on sparsity-based feature selection methods, because sparsity is motivated by the prior knowledge that the most discriminative features are only a small portion of the whole set of features in most medical image analysis problems. While sparsity is considered as a "blessing of dimensionality", sparsity alone is not sufficient for making reasonable and stable inferences. Plain sparse learning models such as $\ell_1$-norm regularized models (Lasso, for example [6]) often provide overly sparse and hard-to-interpret solutions where the selected features are often scattered. Specifically, if there is a set of highly correlated features, then only a small portion of representative discriminate features are selected, resulting into a large false negative rate and a potential biomarker that is hard to trust and interpret. In other words, while the resulted classifier might still achieve high classification accuracy and also be concise, the discovered "biomarkers" might be incomplete and not reliable enough for interpretability.

In particular, for neuroimage analysis, there in general exists intrinsic spatial feature correlations. For example, the measure at a voxel is correlated with measures from its neighbors and this phenomena have been widely observed. When performing feature selection, most existing works focus on select a minimum set of discriminative features and eliminate those redundant though also discriminative features. This is beneficial if the main purpose is to establish a concise and simple classifier. However, if the purpose mainly is to discover the potential interpretable "biomarkers", the plain sparsity based feature selection methods are far from enough, because they are unlikely to select these redundant features. Moreover, we will demonstrate that with the help of reduendant but informative features, the resulted classifier will be more robust in many cases.

To make use of the feature correlation, the elastic net [7] method tries to consider the "grouped influence" by adding an $\ell_2$ regularization to the traditional $\ell_1$-norm penalty to establish a network. The $\ell_2$-norm regularization encourages a grouping effect, where strongly correlated features tend to be in or out of the model together. Elastic net has been used in medical imaging analysis such as MRI reconstruction [8], localizing tumors in images of combined X-ray computed tomography and positron emission tomography [9], Mass Spectrometry Imaging Processing [10]. However, like most existing research, the elastic net regression method mainly focus on fitting a model and cares more about the prediction accuracy. About the feature selections, it lacks a scheme to control the false positives and false negatives.

Recently, there has been several efforts in the finite control of false discoveries of variable selection in the statistical community. For example, stability selection [11] is an important class of methods for high dimensional data analysis with the finite control of false positives. As a special "ensemble learning" procedure, stability selection is an effective approach to stably and reliably perform feature selection and structure estimation based on subsampling on the training samples [11].

While stability selection is a feasible way for the control of false positives, its performance in terms of sensitivity, i.e. the control of the false negatives might be bad, because the traditional stability selection uses the $\ell_1$-norm regularized models as the baseline models, especially when the computational cost is limited and the iterations of stability selection cannot be huge.

In this paper, we consider a less studied topic—stable and comprehensive feature selection with the purpose to discover the more interpretable potential biomarkers, instead of the commonly focused classification accuracy. We aim to propose a novel algorithm to increase the sensitivity performance of the traditional stability selection in the existence of correlated or redundant features. The better performance will be validated by the both synthetic problems and real problems. Notice that, the main benefit of the stable and more comprehensive discriminative feature selection is to discover a better and more reliable interpretable potential biomarkers. The meanings of biomarkers are often field-specific. Below we first briefly review several commonly used feature selection methods in fMRI data analysis beyond sparsity related methods.

*B. Brief Review of Feature Selection Methods for fMRI data*

Typical methods of feature selection for fMRI data are univariate feature selection strategies [12] such as t-test, analysis of variable (anova) and Pearson correlation using simple univariate statistical parameters (e.g., average, variation and correlation coefficient). They are directly testable, easily interpretable, and computationally tractable. Selecting subsets of features as a pre-processing step is independent of the chosen predictor. However, recent studies have demonstrated that "mental representations" may be embedded in a distributed neural population code captured by the activity pattern across multiple voxels [13-15]. In addition, we will demonstrate that two sample t-test method is not robust in case of wrong labels in this paper. Thus, univariate method may not be enough for fMRI feature selection.

Most existing multivariate feature selection methods for fMRI data, are also called multi-voxel pattern analysis (MVPA), if the voxels are considered as features. MVPA is an emerging approach that applies a decoding scheme to all voxels in the entire brain volume simultaneously. The MVPA has proven to be highly useful to decode different patterns of brain activities [14,16]. Most existing multivariate methods such as support vector machine (SVM) and logistic regression work well for the task of classification or prediction. However, the feature selection results based on the obtained weights of voxels during the training of SVM or logistic regression models fail to provide stable and reliable feature selection results, especially when correlated and redundant features exist [17], though the resultant classifier might still achieve satisfying classification accuracy.

Among MVPA, the sparsity regularized models are being widely adopted in the field of fMRI data analysis, due to the same challenge of high dimensional feature space vs. relatively few samples [18]. As mentioned above, however, sparsity alone is insufficient to make interpretable and stable inferences of the discriminative features, especially when there are many discriminative features that are highly correlated to each other. Correspondingly, ones aim to make use of the fact that the active voxels are often grouped together in a few clusters [19]. Beyond the elastic net method, ones also propose to use structured sparsity models [57] by enforcing more structured constraints on the solution. For example, the discriminative voxels are grouped together into few clusters, where the (possibly overlapping) groups have often been known as a prior information [58]. However, in practice, the explicit grouping information is often not available beforehand.

Note that all the above multivariate feature section methods are lack of the guaranteed control of false discoveries of feature selection results. Therefore, stability selection [11] is also being applied in some fMRI studies now and achieved better results than classic plain $\ell_1$ models [20, 21]. In this paper, we will adopt a recent implementation of stability selection which will be reviewed in Section 2.

*C. Our Contributions*

In this paper, we aim to achieve accurate and reliable feature selection for neuroimaging data analysis in order to achieve better interpretability of the discovered potential biomarkers. We proposed a novel feature selection method combining a very recent implementation of stability selection and the elastic net model in order to achieve the better control of both false discovery rate and missed discovery rate. Under the framework of the stability selection, our chosen baseline model is elastic net, rather than the plain $\ell_1$ model, in order to reliably select those informative but redundant features. The more complete and interpretable potential biomarkers are revealed, and the scientific meanings of them are also discussed. Moreover, we are among the first to demonstrate the "unrobustness" of the widely used two-sample t-test method and show the "robustness" benefited from the incorporation of stability selection. Specifically, we designed several experiments to show the "robustness" of our feature selection results, based on a simulation data with noisy labels, as well as the robustness of the resulted classifier based on those redundant but informative features, demonstrated by a multi-center attention-deficit/hyperactivity disorder (ADHD) fMRI data, which is of big variation across different centers. Note that in this paper, we not only deal with the raw fMRI data, but also brain networks which are considered as an example of feature extraction of fMRI. We aim to show that the proposed algorithm is not only applicable to the voxel feature space (raw data), but also some extracted feature space based on fMRI data.

The organization of this paper is as follows. In section II, we first briefly review the stability selection and its recent variant, as well as elastic net methods for feature selection, respectively. Then our method based on them are proposed. In section III, we give the detailed description of the neuroimaging experimental settings. In section IV, the results of our feature selection method on both simulation data and real fMRI data including a multi-center affective disorder (ADHD) are given, compared with other state-of-the-art alternatives. In section V, a short summary of our work and some possible future directions are discussed.

## II. ALGORITHMIC FRAMEWORK

Feature selection originates in machine learning and statistics. In general, redundant features are often considered not to provide additional information, even they are relevant features, if the classification or prediction accuracy is the main goal. Traditionally, the purpose of feature selection is to reduce the risk of overfitting and, hence improve the generalization performance of the resulted classifier as well as reduce the disturbance of the data noise on the classification accuracy. In this paper, feature selection is not only for avoiding over-fitting and generating a concise and reliable classifier, but also to help discover the potential biomarkers for possible diagnosis of affective disorder or certain affective responses. Therefore, the redundant features if they are also informative, need to be selected for better interpretation of the discovered biomarkers, i.e. both the control of false positives and false negatives need to be considered. In addition, as [22] pointed out, these redundant but informative features can help enhance the robustness of the resulted classifier and we verified this point via the multi-center fMRI data analysis.

For neuroimaging analysis, the phenomenon of high-dimensional feature space with few training examples is quite common. We need to make use of certain prior knowledge to help achieve reliable feature selection and reduce the variance of the results, for example, one prior knowledge is the only small portion of features are the discriminative features among all of them. In this paper, we mainly focus on the embedded methods for feature selection, because they are easier to incorporate prior knowledge via the adoption of the regularization terms, than other alternatives such as filters methods (e.g. Chi squared test). Examples of regularization based embedded methods are the Lasso [6], elastic Net and Ridge Regression, which will be briefly reviewed below.

*A. Elastic Net*

In this paper, we adopted the widely used supervise learning method to select the most important features from the given labeled training data. Linear models or Ensemble of linear models have been proved to be sufficient to produce effective classifiers for fMRI data because of high-dimensionality of feature space and relatively small number of training samples [19].

$$Y = Xw + \epsilon \quad (1)$$

where $Y \in \mathbb{R}^n$ is the binary classification label information, so that $Y_i \in \{0,1\}$. $X \in \mathbb{R}^{n \times p}$ is the given training fMRI data and $w \in \mathbb{R}^{p \times 1}$ is the unknown weight reflecting the importance of each feature and is the main basis of feature selection. Two common hypotheses have been made for neuroimaging data analysis: sparsity and grouping effect. Sparsity means that few relevant and highly discriminative voxels are implied in the classification task; by grouping effect, we means feature correlations. For example, in the raw neuroimaging imaging data, there exists ``compact structure'' where relevant discriminative voxels are grouped into several distributed clusters, and are strongly correlated inwardly. In a more general way, the grouping effect does not necessarily mean that these correlated features are neighbored or spatially contiguous, though it might be common when we take the voxels of the raw neuroimaging data as the features. For example, when we use the connections of between different regions as the features, some of these features of high correlations are not necessarily neighbored connections. In short, these two hypotheses need to be made use of when feature selection algorithms are designed.

Elastic net [7] is based on a hybrid of $\ell_1$ regularization and $\ell_2$ regularization and is applied to linear models here. The corresponding objective function for feature selection is written as follows:

$$\min \|y - wx\|^2 + \lambda_1 |w| + \lambda_2 \|w\|^2 \quad (2)$$

where $\lambda_1 > 0$ is the parameter of $\ell_1$ regularization; and $\lambda_2 > 0$ is the parameter of $\ell_2$ regularization. Elastic net can select the relevant voxels by counting the nonzero coefficients of $w$. As mentioned before, the elastic net encourages a grouping effect, where strongly correlated predictors tend to be in or out of the model together. It is particularly useful when the number of features ($p$) is much more than the number of samples ($n$) as is shown in fMRI data. If only $\ell_2$ regularization is adopted, than we have a Ridge regression model. We can adopt other kinds of loss functions besides the least squares residuals such as logistic loss functions.

However, while elastic net respects these two hypotheses of fMRI data, elastic net based feature selection fails to provide a stable feature selection result, i.e. lack of finite sample control of false discovery rate, just like most existing multivariate feature selection methods. Among some recent efforts, stability selection is an efficient way toward the effective control of false variable selection in the plain $\ell_1$ norm regularized linear model (named Lasso [6]) by refitting the Lasso model repeatedly for subsamples of the data, and then keeps only those features that appear consistently in the Lasso model across most of the subsamples [23]. Therefore, we aim to incorporate the idea of stability selection into elastic net in order to achieve an empirical control of false discoveries.

*B. Stability Selection*

We first give a brief review of stability selection. Stability selection is originally proposed in [11] mainly by subsampling of the observations. It considers the variable selection in statistics, also called feature selection in machine learning community. The motivation is that many feature learning selection methods share the same drawbacks of being unstable with respect to small variations of the data, i.e. when one performs feature selection on deferent sets of training data coming from the same source, the results can still very significant. As mentioned above, it is not a big concern when the prediction is the main goal. However, it makes the identification of relevant features quite difficult.

Stability selection, as a method to overcome the above difficulty, consists of applying repeatedly a feature selection method (for example, the plain $\ell_1$ regularized model in the original stability selection [11] to randomly chosen subsamples of the half size of the training samples. The final selection is obtained by picking only those features whose selection frequency across repetitions exceeds a certain threshold. In [24], a variant of stability selection, named complementary pairs stability selection was proposed, but it is still based on the subsampling of the training samples. Stability selection is able to control false discoveries effectively empirically, and gives theoretical guarantees of asymptotically consistent model selection.

*C. Covariate Subsampling*

In [25,26], the authors considered an extension of the original stability selection proposed in (Meinshausen and Bühlmann, 2010). The extension not only considers the subsampling on the observations, but also considers the subsampling on the features. For the given training data matrix $X \in \mathbb{R}^{n \times p}$, this extended stability selection consists of applying the baseline method to random submatrices of X of size $\lfloor n/L \rfloor \times \lfloor p/V \rfloor$, and returning those features having the largest selection frequency. The original stability selection can be roughly considered using a special parameter where L = 2 and V = 1. It has been showed that the bigger L leads to higher independence among different subsamples and results in variance reduction. Feature subsampling ($V > 1$) is conducted to solve the problem of "mutual masking" of relevant features, a problem that happens when relevant feature are intercorrelated. As discussed in [26], the automatic choice of L and V is a thorny theoretical question. In practice, ones have found that the final results are not very sensitive to choice of L>=2, i.e. taking smaller subsample on the training samples does not degrade the final performance. In case of low memory or parallelization, ones can choose a relatively large L. As for V, some initial intuitive analysis shows that in certain circumstances (heavy tailed score noise, for example), restricting the search to a random subset increases the probability of correct recovery [26]. However, the more precise rule for the optimal choice of the subsamples size is yet left for future study. In this paper, L and V are set as 2, and 5, respectively.

To our best knowledge, we are the first to apply subsampling on features, together with the elastic net. The elastic net helps to better take the correlation property of features into consideration and therefore often results into a smaller missed recovery rate on the informative though probably redundant features. The subsampling on features besides on training samples is expected to make the selection of redundant though informative feature more efficient via the reduction of "mutual masking" of relevant features.

*D. Our Feature Selection Algorithmic Framework*

In order to achieve the control of both false discoveries and missed discoveries, we propose to combine the recent implementation of stability selection with the elastic net. As for stability selection and covariate subsampling, the random subsampling avoids the case that certain features only take effect under a fixed combination and therefore improve the generalization of the feature selection result. The paid price is the probably very high false negative rate, i.e. many true relevant features are missed. On the contrary, the elastic net takes the feature correlation (or called grouping effect) into consideration and help reduce the missed discovery rate [7]. While the stability selection has already proved to be able to control false discoveries in practice [11, 26], their combination is expected to further reduce the false negatives.

We first gave an overall description of the algorithmic framework. First, denote the number of resamplings as N. During each resampling step of stability selection, every subsampling random submatrices of the given training data matrix $X \in \mathbb{R}^{n \times p}$ is denoted as submatrices $\tilde{X}_j$ of size $\lfloor n/L \rfloor \times \lfloor p/V \rfloor$. The corresponding label vector is denoted as $y_j \in \mathbb{R}^{\lfloor n/L \rfloor \times 1}$. Let F be the set of indices of all $p$ features, and let $f \in F$ denote a feature. If the feature $f$ is not selected in the submatrix $\tilde{X}_j$, $w_f^{(j)} = 0$. Otherwise, we estimate $w_f^{(j)}$ from the random submatrices $\tilde{X}_j \in \mathbb{R}^{\lfloor n/L \rfloor \times \lfloor p/V \rfloor}$ and $y_j \in \mathbb{R}^{\lfloor n/L \rfloor \times 1}$, based on the baseline model--elastic net. For a feature f, if $w_f^{(j)} \neq 0$ then the feature is considered to be relevant feature. Denote $S(\tilde{X}_{(j)}) = \{f: w_f^{(j)} \neq 0\}$ as the set of features selected based on $w^{(j)} \in \mathbb{R}^{\lfloor p/V \rfloor \times 1}$. The procedure is repeated N times and we can get the stability score for every feature by:

$$SS(f) = \frac{1}{N} \sum_{j=1}^{N} \mathbf{1}\{f \in S(\tilde{X}_{(j)})\} \quad (3)$$

where $\mathbf{1}\{\cdot\}$ is the indicator function.

Finally, given the number of features we desired to be included in the model, we can choose top ranked features by stability score as filters-based feature selection methods do.

We would like to point out that the original stability selection proposed in [11] is mainly on random subsampling of observations, i.e. the rows of X. As the paper by [26] has also pointed out, the random subsampling in terms of observations can in general guarantee the finite control of false positives, even though different base methods are adopted. Therefore, while we are using a more complicated base method elastic net, rather than the plain $\ell_1$ norm regularized model, the finite control of false positives can be still achieved. Moreover, we expect a better empirical performance of control of missed discoveries, benefiting from the incorporation of the correlation of features by the elastic net as well as the subsampling on the subsampling on the features.

The procedure of our algorithm is summarized in the following table.

The Algorithmic Framework of Stable Feature Selection Method

**Inputs**:
(1) Datasets $X \in \mathbb{R}^{n \times p}$
(2) Label or classification information $y \in \mathbb{R}^n$
(3) Elastic net $\ell_1$ regularization parameter $\lambda_1$ and $\ell_2$ regularization parameter $\lambda_2$.
(4) Number of randomizations N, sub-sampling fraction $\alpha \in [0,1]$ in terms of rows of X; sub-sampling fraction $\beta \in [0,1]$ in terms of columns of X
(5) Initialize stability scores: $SS(f) = 0. f \in F$

**Output**: stability scores $SS(f)$ for all $f \in F$

**Procedure:**
For *j=1* to N
(1) Perform sub-sampling in terms of rows: $X \leftarrow X_{[K,:]}, y \leftarrow y\mathcal{L}$ where $\mathcal{L} \subset \{1,2,\dots,n\}$, $card(\mathcal{L}) = \lfloor \alpha n \rfloor$, the updated $X \in \mathbb{R}^{\lfloor \alpha n \rfloor \times p}$ and the updated $y \in \mathbb{R}^{\lfloor \alpha n \rfloor}$.
(2) Perform sub-sampling in terms of columns: $X \leftarrow X_{[:,\daleth]}$, where $\daleth \subset \{1,2,\dots,p\}$, and $card(\daleth) = \lfloor \beta p \rfloor$
(3) Estimate $w^{(j)} \in \mathbb{R}^{\lfloor \beta p \rfloor}$ from X and y with elastic net
(4) Store indices of selected features:
$$S(\tilde{X}_{(j)}) = \{f: w_f^{(j)} \neq 0\}$$
End for
Now, we can compute the stability scores for all f:
$$SS(f) = \frac{1}{N}\sum_{j=1}^{N} 1\{f \in S(\tilde{X}_{(j)})\}$$

*E. Some preliminary rethinking of our algorithms*

As we know, stability selection and its variants belong to the family of the more general ensemble learning methods. However, they focus on feature selection, while most existing ensemble learning algorithms focus more on achieving better classification accuracy. Our algorithm following the idea of the stability selection aims to achieve a feature selection with a controlled false positive rate. Moreover, unlike most of existing feature selection methods, we would like to find out those informative though redundant features, not only for better interpretation of the potential biomarker, but also for achieving a more robust classifier, which is seldom mentioned in existing literatures and will be demonstrated in Section III and IV.

Notice that for neuroimaging data analysis, directly considering the raw data and taking voxels as the features is only one way to perform pattern recognition and classification. There have existed a lot of efforts to exact more appropriate feature extraction of neuroimaging data, depending on different circumstances, because appropriate features are important to achieve effective classification or prediction accuracy, as well as obtain useful potential biomarkers [27, 28, 29, 30].

For some newly extracted features such brain networks based on voxels correlations, correlation between new features is not followed by "spatial contiguousness". So the methods based on "spatial contiguousness" like those proposed in [31,32] may not be applicable. However, the new feature space is still in general high dimensional and the number of samples is typically relatively small, i.e. n<<p for X still holds. In such cases, the correlation between features, i.e. the columns of X still exists and therefore, our proposed algorithm is still applicable, because it does not depend on the "spatial contiguousness". In other words, when considering the feature selection on fMRI data, we not only consider the voxels feature of the raw data, but also consider connections as features where we generate a brain network based on the raw fMRI data.

## III. EXPERIMENTAL SETTINGS

In this study, we developed a novel data-driven feature selection approach by integrating elastic net and an idea of stability selection method. Our results on the synthetic data and real neuroimaging data including multi-center data of affective disorder indicated that the novel integrated approach may be a valuable method for potential biomarker extraction and pattern recognition of fMRI data.

We aim to demonstrate 1) better control of both false discovery and missed discovery of our proposed method. 2) the robustness of feature selections in existence of label noise, 3) We would also like to show the completeness of feature selection of our algorithm helps generate a more robust and accurate classifier via multicenter fMRI data analysis.

As for the point 1), we use both the synthetic data and real fMRI data to verify this point. Then based on the true fMRI data, we will demonstrate the scientific meanings of the better feature selection results in the context of potential neuroscience "biomarker" discovery.

As for point 2), the need of robustness and reliability in feature selection is often amplified by the challenge from the label noise. In this paper, we design a robust test that add some label noise to the simulation/synthetic data. A previous study [33] has proved that label noise is potentially harmful than feature noise, highlighting the importance of dealing with this type of noise. The detailed description of the generating procedure is presented in the following Subsection 3.1.1.

As for the point 3), it is the data noise or data variation. It is often very hard to obtain high quality training data in practice. The form of training data depends on specific tasks and the source data quality. Because of the highly noisy nature and high consumption nature of fMRI data, a robust and stable feature selection method is quite necessary. We demonstrate the advantage of our algorithm in this aspect via the multi-center ADHD data, where data variation (noise) is significant across different centers. We make use of the multi-center ADHD data to show the selection of redundant informative features help construct a more robust classifier. Traditionally, it is a challenge to construct a robust classifier due to the data variation across different centers.

*A. Test Data*

*A.1 Synthetic Data Generation*

In this paper, a synthetic data $(70 \times 63\ pixels)$ was generated on an axial brain as shown in Fig.1 (a). There were five sub-regions with each of them contained $7 \times 7 = 49$ pixels, as shown in the Fig 1(a) in white. The time series of all pixels of a subregion consisted of a certain signal mixed with Gaussian noise under a signal-to-noise ratio (SNR=1.0; SNR is defined as the standard deviation ratio between signal and noise). Three active temporal patterns with three delay versions (delay of 0, 5, 10 time points; see Fig 1(b)) of the "expected" boxcar-like timing function were depicted. Three different active temporal patterns were added to subregion "A", subregions "B", "C", and "D", and subregion "E", respectively (see Fig 1(a)). In the current study, we use time point signals of pixels as features to identify the potential ones which could classify between active time periods and blank time periods. The second active temporal pattern is designed as the discriminative pattern. The active time points are designed as label '1' and inactive time points are designed as label '0' when we identify discriminative features in all pixels, the subregions "B", "C", and "D" would be the discriminative clustered features correspondingly.

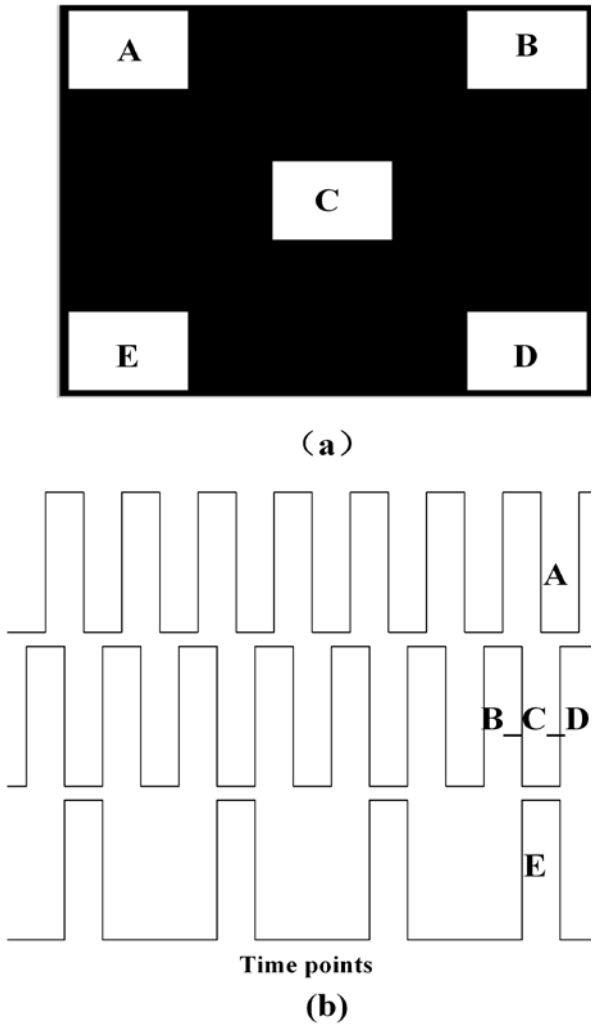

**Fig. 1** Synthetic fMRI image. (a) Spatial distribution of assumed active regions; (b) Three assumed stimulation patterns with alternate ten points at rest and ten points in the task conditions.

*A.2 fMRI Data of Face Recognition Experiments*

Thirty college students participated in this experiment. All subjects were right handed confirmed by the Chinese version of Edinburgh Handedness Questionnaire (coefficients> 40). The subjects had normal or correct to normal vision, and were free from any medications, neurological and psychiatric disorders. The task dataset is a steady-state experimental design which consisted of 30 trials, with each trial comprised of 2s face image stimulation and 18s fixation. Participants were asked to judge whether each face is neutral (right thumb response) or happy (left thumb response). In fact, all faces were neutral. This dataset was compared with a comparable length of resting-state dataset scanned before the task in the same session. One image of brain activity in the dataset is consisting of $61 \times 73 \times 61$ voxels. The data of four subjects were removed from the final analysis due to large head motions (translation >2 mm or rotation >2 degree).

Both resting-state and task data were obtained using a 3.0T GE750 scanner (General Electric, Fairfield, Connecticut, USA) at the University of Electronic Science and Technology of China. The parameters were as follows: repetition time (TR) =2000 ms, echo time (TE) =30 ms, 90 degree flip angle, 43 axial slices (3.2 mm slice thickness without gap), 64*64 matrix, 24 cm field of view.

*A.3 Multi-Center ADHD Data*

Furthermore, we used a multi-center fMRI data to test the performance of our feature selection algorithm. The data were downloaded from the ADHD-200 Consortium for the global competition (http://fcon_1000.projects.nitrc.org/indi/adhd200/). It was acquired in two different sites: Peking University, New York University Child Study Center. There were 62 children, 29 of whom were healthy controls, and the remaining 33 were patients with ADHD in New York University site. There were 74 children, 37 of whom were healthy controls, and the remaining 37 were patients with ADHD in Peking University site. Unlike 3.1.2. Face Recognition fMRI Data, we use the brain connectivities as the features, instead of voxels for this data. More details about how to generate the brain network are given below.

*B. Data-Processing Procedure*

For both Face Recognition fMRI Data and Multi-center ADHD, functional MRI images were preprocessed using the Data Processing Assistant for Resting-state fMRI (DPARSF 2.2, http://restfmri.net/forum/DPARSF) [34] The preprocessing steps included: slice timing; spatial transformation, which included realignment and normalization, performed using three-dimensional rigid body registration for head motion. The realigned images were spatially normalized into a standard stereotaxic space at 2*2*2 mm^3, using the Montreal Neurological Institute (MNI) echo-planar imaging (EPI) template. A spatial smoothing filter was employed for each brain's three-dimensional volume by an isotropic Gaussian kernel (FWHM=8 mm) to increase the MR signal-to-noise ratio. Then, for the fMRI time series of the task condition, a high-pass filter with a cut-off of 1/128 Hz was used to remove low-frequency noise.

For multi-center ADHD data, we consider the brain networks and thus proceed the following pre-processing. Each subject of multi-center fMRI data was further divided into 90 anatomical regions of interests (ROIs) [35] (45 in each hemisphere) according to the automated anatomical labeling (AAL) atlas [36], after that, a representative time series in each region was obtained by averaging the fMRI time series of all voxels in each of the 90 regions by DPARSF software. These representative time series were temporally bandpass filtered (0.01-0.08 Hz), and several sources of spurious variance were removed by regression along with their first derivatives, such as six head motion parameters, white matter signal and cerebrospinal fluid signal. Functional connectivity between each pair of regions was evaluated using Pearson correlation coefficients, resulting in 4005 dimensional functional connectivity feature vectors for each subject. These functional connections are the features used in pattern recognition.

*C. The Alternative Methods for Comparison*

In this paper, we compared our algorithm with the classical univariate voxel selection method, and multi-voxel pattern

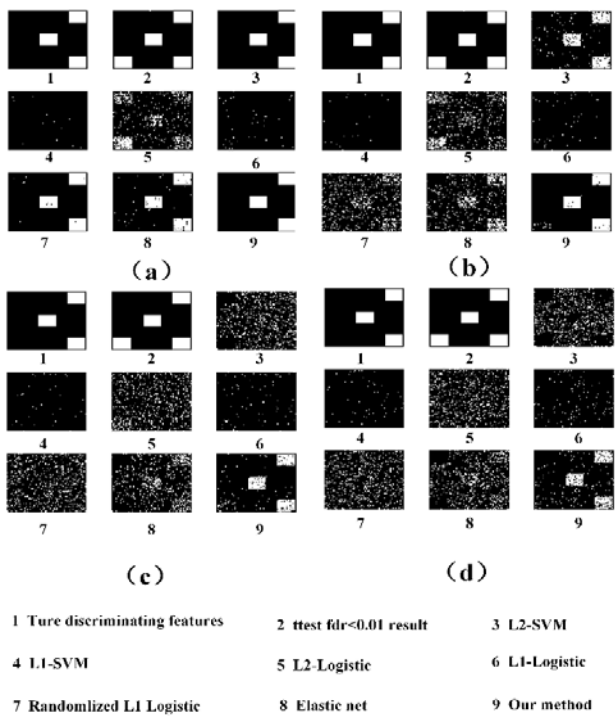

1 Ture discriminating features  2 ttest fdr<0.01 result  3 L2-SVM
4 L1-SVM                        5 L2-Logistic           6 L1-Logistic
7 Randomlized L1 Logistic       8 Elastic net           9 Our method

**Fig. 2** The maps of estimated discriminative voxels by different methods on the synthetic data. (a) the maps of all labels are ture. (b) the maps of one label are wrong. (c) the maps of five label are wrong. (d) the maps of ten label are wrong.

recognition methods, including T-test, $\ell_2$-SVM, $\ell_1$-SVM, $\ell_2$ Logistic Regression, $\ell_1$Logistic Regression, randomized $\ell_1$ Logistic Regression and the Elastic Net. Here randomized $\ell_1$ logistic regression is based on the recent idea of stability selection [25, 26] and has random sampling on the features.

The T-test is implemented as an internal function in MATLAB. $\ell_2$SVM, $\ell_1$SVM, $\ell_2$Logistic Regression, $\ell_1$ Logistic Regression, Elastic Net, have been implemented in LIBLINEAR [37], or SLEP (Sparse Learning with Efficient Projections) software [38]. Randomized logistic regression and our propose algorithms are written based on the available $\ell_1$logistic regression code and the elastic net code.

*D. Parameter Settings of Involved Algorithms*

In this paper, for the regularization parameters of each alternative method for comparison, their choices are mostly based on cross validation unless specified otherwise.

For our proposed algorithm, the baseline model--elastic net is assumed to provide an initial selection procedure in subsampling, thus we hope that the values of the sparse parameters should not be too large, or very few features are selected in each iteration; nor should they be very small, in which case the selection probability will be very high for all the features. Stability selection can make the choice of regularization parameters for both randomized $\ell_1$ logistic and our proposed algorithms is less sensitive to the final results. Therefore, the regularization parameters $\lambda_1$ and $\lambda_2$ of the elastic net sub-problem of each iteration are fixed and the same as those values of the compared alternative: one-time elastic net where all the data without subsampling is used and the parameters are determined based on cross-validation as mentioned in the above paragraph.

For both randomized $\ell_1$ logistic regression and our algorithm, we set the subsampling rate α = 0.5 and β = 0.2. For our algorithm, we set the total resamplings N=200. The resampling number of random $\ell_1$ logistic regression is 500 in all of our experiments. The choice of the number of resamplings is only empirical here.

## IV. EXPERIMENTAL RESULTS and ANALYSIS

A. Simulation Test

In this section, we proposed a label noise robust test by randomly selecting 1-10 sample(s) to reverse their labels (when the selected sample label is '1', turn it to '0', and vice versa), then calculating the discriminative voxels with different pattern recognition methods, to test the robust performance of each involved method on this condition where the training samples have certain number of wrong labeling.

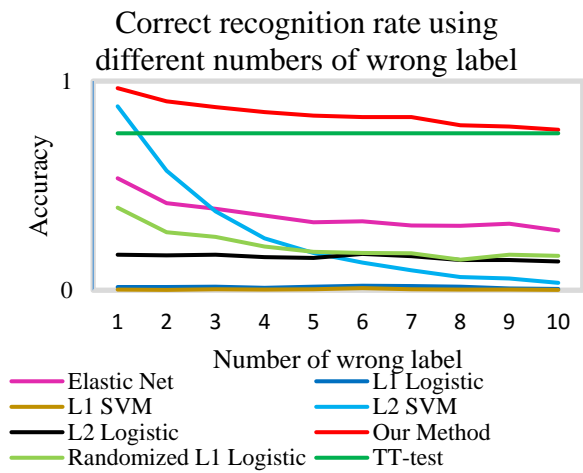

**Fig. 3** Voxel selection accuracy as a function of the number of wrong labels

Fig 2 shows the results of "robustness" test. The Fig 2(a) is the maps of estimated discriminative voxels by different methods on the synthetic data (unthreshold, i.e. the gray level is based on the absolute value of **w**) when all labels are true: our method together with L2-SVM are the only two methods which can find out the accurately discriminative regions.

Figs 2(b)-2(d) are the maps of estimated discriminative voxels when parts of labels are wrong. We can see from Fig.2 that the univariate method (two sample t-test) finds out regions "B", "C",

"D" and "E", as shown in the second subplot, but the accurately discriminative regions are just region "B", "C", and "D", as shown in the first subplot. It is not too hard to understand that two-sample t-test is based on the means of two variables or distinct groups, and the two groups (divided by two time series as shown in Fig.1 (b)) are obviously different in region "E". That is to say, the result of t-test might have false positives. L2-SVM slightly surprises us in this case, because it can find out the true discriminative regions when all labels are true. However, with the number of wrong label increases, its result becomes disorderly and unsystematic. As for the L1-logistic regression and L1-SVM, both of them return over-sparse solutions (Fig.5 displays the same conclusion), which are hard to discriminate and interpret, as expected. The single elastic net is able to approximately find out the right regions when all labels are true, but it has the same problem with L2-SVM that their functions are excessively relied on the quality of data. As for the randomized L1-logistic, the classical stability selection method, it cannot return a satisfying result, especially when some labels are wrong. The results showed that our method has a better robust performance than other methods in case of the existence of wrong labels.

The results in Fig. 3 show that as the number of wrong label increased, the change of selected features by our methods was very small. Even when ten labels are wrong, our method can approximately find out all those true discriminative regions, indicating that our method has a good robust characteristic in terms of feature selection. An intuitive explanation is that subsampling procedure can provide a stable feature section solution, just as an ensemble of classifiers provide enhanced classification performance [19].

The results in Fig.4 show the Precision-Recall Curve of each method when five labels are wrong. Precision is the fraction of retrieved instances that are relevant, while recall is the fraction of relevant instances that are retrieved. While still keeping good control of false positives, our method is the most sensitive.

### B.  Actual fMRI Experiment Test: Face Data

During the fMRI scanning, subjects were under the two conditions: resting-state and face stimuli. Each condition was lasted for 10 min. According to the cardinal haemodynamic response function (HRF), the blood oxygen level dependent (BOLD) response should be the strongest at the $3^{rd}$ and $4^{th}$ time points, so the data we used here was the mean of the $3^{rd}$ and $4^{th}$ time points data. Then, the number of samples of each subject is 60, in which 30 samples are for resting-state, and the other 30 are for face stimuli state, respectively. We used an averaged data based on all the 26 subjects.

We anchored five regions from visual processing (the right occipital face area (OFA), the right fusiform face area (FFA), the right posterior superior temporal gyrus (pSTG)) to motor action (the supplementary motor area (SMA), and left sensorimotor cortex (SMC)) to describe the time course of face recognition. The OFA, FFA, and pSTG are core regions of face recognition [40]. The OFA is thought to be involved in the early perception of facial features and has a feed-forward projection to both the pSTG and the FFA, the connection between the OFA and pSTG is thought to be important in processing dynamic changes in the face [40]. It has been suggested that the SMA could be implicated in facial emotion expression and recognition [45], activity in the sensorimotor areas serves as a marker of correctly recognizing emotional faces [46]. The OFA and FFA were well captured by our approach, elastic net, L2 SVM and L2 Logistic (see Figure 5 at the very end of the manuscript, the third column of each approach); the pSTG was obtained by our method, L2 SVM, L2 Logistic and Rand-L1,(Figure 5, the fourth column of each approach); the SMA was detected by our method, elastic net, L2 SVM, L2 Logistic and Rand-L1 (Figure 5, the second column of each approach); and the SMC was captured by our method, elastic net, L2 SVM and L2 Logistic (Figure 5, the first column of each approach). Therefore, three approaches including our method, L2 SVM and L2 Logistic, can reveal all these five regions. However, only our method obtained the most complete and spatially continuous regions, resulting into the most distinguishing results. Furthermore, our approach can detect more clustered regions than other methods, which are in line with opinions that steady-state brain responses have high signal-to-noise ratio (SNR) [41-43] and most of brain regions should respond to cognitive tasks when the SNR is high [44]. Notice that L1-logisitc and L1-SVM reveals very few voxels as expected because the pure sparsity regularization is likely to result in very few scattered selected voxels, especially when the number of samples is also small.

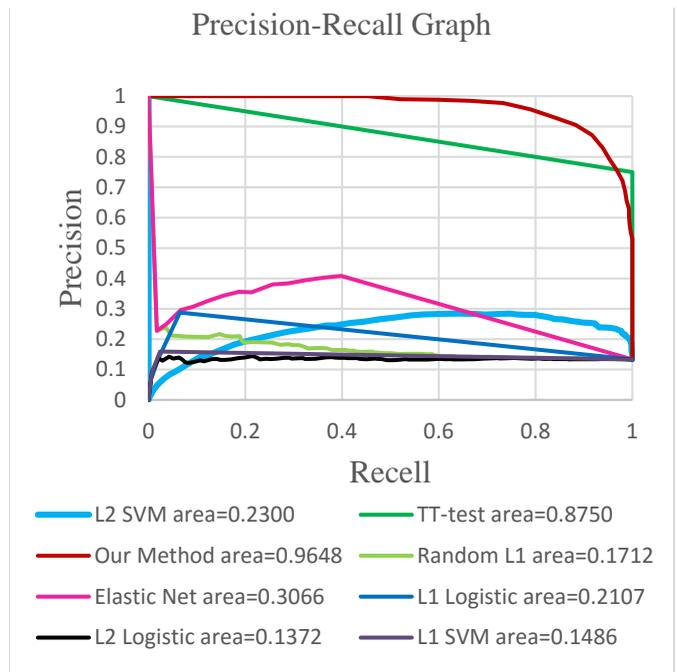

Fig.4 The precision recall curve when five labels are wrong

In summary, Fig.5 (at the very end of the manuscript) showed that our method can detect the five regions involved in the time course of facial recognition, including OFA, FFA, pSTG, SMA and SMC. The five regions are the core regions in the face recognition stream from visual information processing to motor output. Current results indicate that our method is better at detecting key features in cognitive activities than other alternative approaches. Specifically, it can achieve both better false discovery control and missed discover control, and this advantage is quite important for revealing the meaningful biomarkers for either medical diagnosis or cognitive study.

## C. Actual fMRI Experiment Test: Multi-Center ADHD data

*C.1 Feature Selection*

We first applied our feature selection method to the data of Peking University. After calculating the score of each feature (edge of brain network), the weight of each region (node of the brain network) could be evaluated by summing one-half of the feature scores associated with that region [47] to represent the relative contributions of different regions. Some regions showed greater weights than others. Specifically, we defined a region having significantly higher weight if its weight was at least one standard deviation greater than the average of the weight of all regions, as did in previous studies [48,49]. The regions with the higher weights included the left precentral gyrus (PreCG), right superior frontal gyrus (SFG), right rolandic operculum (ROL), left olfactory cortex (OLF), left anterior cingulate cortex (ACC), left meddle cingulate cortex (MCC), left lingual gyrus (LING), right inferior occipital gyrus (IOG), left superior occipital gyrus (SOG), bilateral fusiform gyrus (FG), left inferior parietal lobe (IPL), left supramarginal gyrus (SMG), e right angular gyrus (ANG), and right temporal pole (TP). The region of the left IPL exhibited the highest weight. Fig. 6 displays these regions.

From Fig.6, we can see the regions with high weights were related with the default mode network (DMN:IPL,ANG), the ventral attention network (VAN: ROL/VFC, SMG, ANG/TPJ), the dorsal attention network (DAN: PreCG/FEF, ANG/IPS), executive control network (ECN: SFG/dlPFC, ANG, IPL/PPC) and the visual network (VN: IOG, SOG, LING). A recent study pointed the altered resting state functional connectivity of ADHD between the DMN and VAN (ventral attention networks) [50]. The DAN (dorsal attention network), ECN (executive control network) and VN (the visual network) also have been found to be affected by the methylphenidate, a primary treatment for ADHD [51]. More importantly, these five networks have been reported to be malfunctioned in patients with ADHD by studies using the ADHD-200 dataset [54,55]. The high consistency among studies indicates that our method can successfully detect core networks that are abnormal in ADHD. These results, therefore, demonstrated the effective of our method in selecting true positive features and rejecting false negative features in real resting state fMRI data.

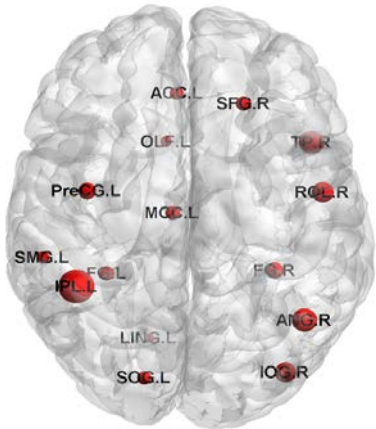

**Fig. 6** Rendering plot of the regions with significantly higher weight in the classification. The size of the node represented the magnitude of the normalized region weight. L left. R right

*C.2 Classification Accuracy Tested on Data of Another Center*

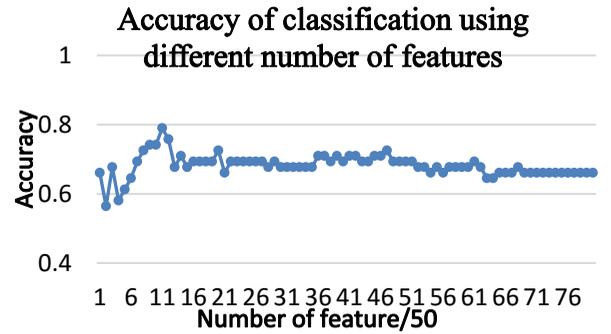

**Fig. 7** Predictive accuracy as a function of the number of features used in the classification process. The features were ranked according to stability score in descending order.

After using the data of Peking University as training data to rank the features by our method, the data of New York University was used as test data with a leave-one-out cross-validation (LOOCV) strategy to evaluate the performance of a classifier (linear-svm). This is among the first efforts to perform cross val-idation based on different centers. Many existing multi-center works put the data from different centers into one pool and perform leave one out validation [56].

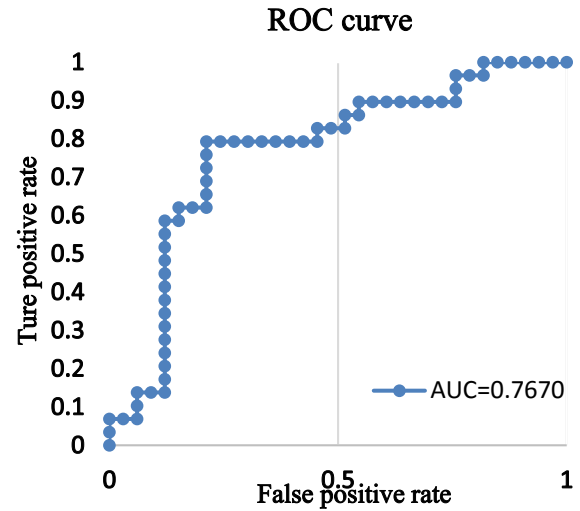

**Fig. 8** ROC curve of the classifier. ROC receiver operating characteristic

As show in Fig.7, the classifier could reach up to 79.03% (76.67% for sensitivity, 81.25% for specificity) by using the top 550 highest ranked feature. Taking each subject's discriminative score as a threshold, the receiver operating characteristic (ROC) curve of the classifier was yielded, as shown in Fig. 8. The area under the ROC curve (AUC) of the proposed method was 0.7670, indicating a good classification power. Since the fMRI data collected from different centers may have some systematic differences that are possibly caused by the different types of MRI machines and settings, our method shows a stable and robust result in case of data variation [35,52].

*C.3 Comparing with Other Algorithms*

In this section, the same procedure is applied to the alternative methods include two-sample t-test, randomized $\ell_1$ logistic, $\ell_2$ logistic, $\ell_1$ logistic, $\ell_2$ SVM, $\ell_1$ SVM and Elastic net. The results are showed in Fig. 9 and Table 1.

Fig.9 shows how the predictive accuracy varies with the number of most relevant features used in the classification process. The horizontal axis represents the value of the number of selected features divided by 50. The L2-svm achieves the best accuracy of 67.74% (66.67% for sensitivity, 68.57% for specificity) when the 200 highest ranked features are used; Randomized L1-logistic achieves the best accuracy of 67.74% (64.52% for sensitivity, 70.79% for specificity) when the 450 highest ranked features are used; Elastic net achieves the best accuracy of 72.58% (68.75% for sensitivity, 76.67% for specificity) when the 1300 highest ranked features are used; two sample t-test achieves the best accuracy of 77.42% (72.73% for sensitivity, 82.76% for specificity) when the 50 highest ranked features are used; L1 logistic achieves the best accuracy of 70.97% (68.97% for sensitivity, 72.73% for specificity) when the 350 highest ranked features are used; L2 logistic achieves the best accuracy of 72.58% (71.47% for sensitivity, 73.53% for specificity) when the 600 highest ranked features are used; L1 SVM achieves the best accuracy of 69.53% (66.67% for sensitivity, 71.88% for specificity) when the 500 highest ranked features are used. These highest classification performance corresponding to different feature selection methods are listed in Table 1. The corresponding sensitivity and specificity are also listed. It shows that our method performs better than other methods in terms of not only in accuracy, but also in sensitivity, and specificity in terms of classification accuracy. To summarize, our method has demonstrated to be effective, and has a better robust performance than other methods here by selecting those redundant but informative features, especially in the multi-center case where the data is of big variation across different centers.

We have shown the "robustness" of our method in terms of feature selection when there exist noisy or wrong labels in the first numerical experiments based on synthetic data. This multi-center data based experiment further demonstrates that the accuracy and completeness of feature selection can also help generate a more robust and accurate classifier in the existence of data variation. This phenomenon accords with other related studies such as [53], where they also claim the comprehensive feature selection enhances the robustness of the resultant classifier.

## V. SUMMARY and DISCUSSION

In this paper, we introduced a stable feature selection method which combines stability selection and elastic net for fMRI data, which often has correlative and redundant features of high dimensionality. We tested the effectiveness of this algorithm on a synthetic dataset and two real fMRI datasets, especially for a multi-center data of mental disorder (ADHD). The results indicated that this algorithm could effectively select discriminative features for high dimensional data with a better empirical control of both false positives and negatives. These results suggest that our method be suitable in revealing potential biomarkers than other alternative approaches. The more accurate, complete and robust discovering of true discriminative features provides a sound support for the scientific research of neuroscience to a certain degree. In addition, the classifier based on feature selection results of our algorithm can achieve a superior prediction accuracy with robustness, which is demonstrated by a multi-center data analysis for the first time to our best knowledge.

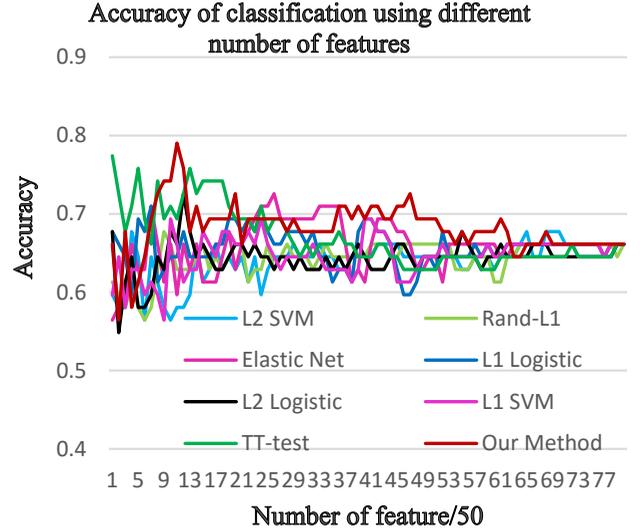

**Fig. 9** Predictive accuracy as a function of the number of features used in the classification process by using linear SVM. The features were ranked according to different weights descending order descending order

**Table. 1**
Classification performance of different feature selection methods.

| Method | Accuracy (%) | Sensitivity (%) | Specificity (%) |
|---|---|---|---|
| L2-SVM | 67.74 | 66.67 | 68.57 |
| Randomized L1 Logistic | 67.74 | 64.52 | 70.97 |
| Elastic Net | 72.58 | 68.75 | 76.67 |
| TT-test | 77.42 | 72.73 | 82.76 |
| L1-Logistic | 70.97 | 68.97 | 72.73 |
| L2-Logistic | 72.58 | 71.43 | 73.53 |
| L1-SVM | 69.35 | 66.67 | 71.88 |
| Our Method | 79.03 | 76.67 | 81.25 |

However, the results are mostly empirical and we might need to perform theoretical support in terms of possible bounds of the false positive rate and especially false negative rate in the future work. Furthermore, how to effectively distinguish true positives from false positives needs to be better addressed. In addition, the robustness to the data variation and label noise might need a theoretical analysis, though it has been empirically demonstrated in this paper.

**Acknowledgment**
This work was supported by 863 project (2015AA020505),

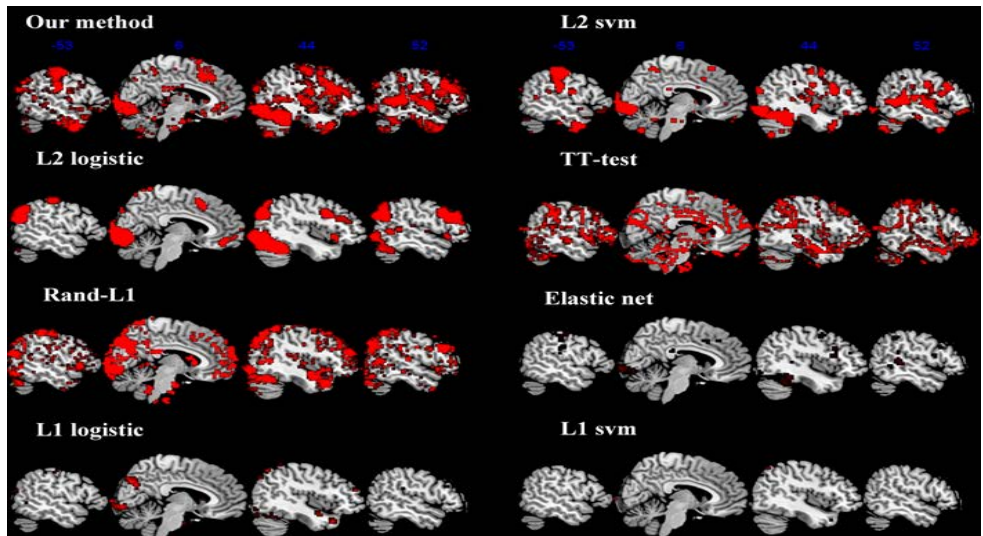

**Figure 5**: Score maps estimated by different methods for fMRI Data of Face Recognition Experiments